\documentclass{article}

\usepackage{iclr2027_conference,times}

\usepackage{amsmath,amsfonts,bm}

\def\eqref#1{equation~\ref{#1}}

\def\1{\bm{1}}

\DeclareMathAlphabet{\mathsfit}{\encodingdefault}{\sfdefault}{m}{sl}
\SetMathAlphabet{\mathsfit}{bold}{\encodingdefault}{\sfdefault}{bx}{n}

\usepackage{amsmath}
\usepackage{amssymb}
\usepackage{booktabs}
\usepackage{multirow}
\usepackage{array}
\usepackage{float}
\usepackage{graphicx}
\usepackage{xcolor}
\usepackage{microtype}
\usepackage{hyperref}
\hypersetup{hidelinks}
\usepackage{url}

\newcommand{\mht}{\textsc{MHT}}
\newcommand{\fine}{Concurrent Functional Loading}
\newcommand{\coarse}{Serial Functional Scheduling}
\newcommand{\base}{Baseline}
\title{Can Large Language Models ``Hyper-Thread''?}

\author{
Fei Ding\thanks{Corresponding author: \texttt{dignfei@gmail.com}}\\
Alibaba Group
}

\iclrfinalcopy

\begin{document}

\maketitle
\fancyhead{}

\begin{abstract}
Large language models generate tokens sequentially, but can they execute multiple tasks concurrently while forming each token? Broader attention allocation may provide a mechanism for such task concurrency. Existing approaches to scaling inference primarily rely on longer generations, more samples, or additional verification stages, while attention dispersion is often treated as a signal of interference or error. Task concurrency within serial generation therefore remains underexplored. We propose the \textbf{Model Hyper-Threading Hypothesis} and evaluate its predictions using multiple coordinated tasks that share state within the same problem. We design three conditions---\base{}, \coarse{}, and \fine{}---and evaluate their benefits and costs using accuracy, output-token distributions, and attention metrics. On an AIME 2025 development set, \fine{} achieves the highest accuracy. Relative to \coarse{}, its typical output length is similar and it is shorter on most problems, while exhibiting greater attention dispersion and higher task-relevant coverage, albeit with a heavier output-length tail. Within-step concurrency and its causal mechanism still require direct tests. Our results show that more dispersed attention can coexist with higher accuracy, providing preliminary behavioral and correlational evidence for the hyper-threading hypothesis. These findings motivate a shift in perspective on inference scaling from ``generating more tokens'' toward ``having each generation step carry more tasks,'' pointing to a new avenue for improving reasoning performance.
\end{abstract}

\section{Introduction}

Prior work often associates some forms of attention dispersion with interference from irrelevant information and reasoning errors. When a model allocates weight to weakly related steps, widespread nonzero connections may introduce interference \citep{bai-etal-2025-self}. At particular stages when answer tokens attend back to a reasoning trace, incorrect answers also tend to exhibit diffuse and irregular attention patterns, whereas correct answers consistently focus on key semantic anchors \citep{chen-etal-2026-answer}. Yet does this mean that dispersed attention is necessarily harmful? We ask a counterintuitive question: when a model has sufficient reasoning ability, could more dispersed attention allow it to access information required by multiple tasks within a single generation step and advance the corresponding functional computations concurrently, thereby improving reasoning efficiency per generation step and final accuracy?

To address this question, we propose the \textbf{Model Hyper-Threading Hypothesis (\mht)}: \emph{serial autoregressive output does not necessarily imply serial functional computation; more dispersed attention may enable multiple interrelated functional computations to execute concurrently within a single generation step, increasing the functional capacity of each step and improving task performance without a disproportionate increase in explicit generation.} Here, ``hyper-threading'' is a hypothesis about functional computational multiplexing, not a claim that a Transformer contains physical threads analogous to those in a CPU.

Autoregressive large language models expose a strictly serial interface: the model can generate the next token only after generating the current one. A reasoning problem, however, commonly requires the model to select a solution path, maintain problem constraints, check boundary cases, verify intermediate calculations, and control the final output at the same time. Existing inference-scaling methods mostly realize these functions by generating longer chains of thought, drawing more samples, or introducing separate reflection stages \citep{NEURIPS2022_9d560961,wang2023selfconsistency,ICLR2025_1b623663,ICLR2024_aca97732}. This encourages an implicit assumption that adding functional load must consume a corresponding number of additional generation steps. The Model Hyper-Threading Hypothesis asks whether serial generation truly implies that functional computation must also proceed serially in separate stages.

As a first controlled test of this general hypothesis, we use solving, verification, constraint tracking, and output control within the same problem as our experimental instance. These functions share inputs and intermediate states and jointly serve the final answer, allowing final accuracy to measure the net benefit of the functional bundle. Specifically, we load accuracy-oriented obligations for solving, logical verification, boundary checking, derivational closure checking, and computational verification. \base{} specifies no additional functional schedule. \coarse{} and \fine{} share verification objectives and error categories, with their prompt differences concentrated in scheduling granularity and its operationalization: the former performs grouped checks after each paragraph, whereas the latter jointly maintains these objectives throughout generation and repairs errors immediately. ``Concurrent'' and ``serial'' here describe only the functional organization imposed by the external input instruction; they do not presuppose that concurrent computation occurs internally. The output-token distribution tests whether accuracy gains require explicit generation to grow proportionally with functional load.

The complete results appear in Tables~\ref{tab:main}, \ref{tab:paired}, and \ref{tab:tail}. \fine{} attains the highest accuracy. Its typical output length is close to that of \coarse{}, and it does not produce longer outputs on most paired problems. Its mean output length and tail risk are nevertheless higher, revealing that \fine{} may amplify both effective recovery and an unproductive search tail.

These results remain exploratory. Each problem has only one generation per condition. \fine{} and \coarse{} form a goal-matched prompt pair designed to minimize differences, but scheduling granularity and immediate repair are not factorially separated. We separately analyze candidate probabilities for visible output tokens and directly measured Transformer self-attention matrices; predictive output entropy and self-attention metrics have different meanings and cannot substitute for one another. The measured attention results show that higher accuracy coexists with attention that is more dispersed and has higher task-relevant coverage, but this does not amount to directly observing or causally establishing within-step functional concurrency. The current evidence is behavioral and correlational: switching from grouped checks after each paragraph to continuous joint loading with immediate repair co-occurs with higher final accuracy, while output length does not increase on every problem.

Our contributions are as follows:
\begin{itemize}
    \item \textbf{A general hypothesis.} We propose the Model Hyper-Threading Hypothesis, distinguish serial token output from serial functional computation, and suggest that more dispersed attention may serve as a computational medium for concurrent execution and multiplexing of related functions rather than as a fixed indicator of failure.
    \item \textbf{A within-problem experimental instance.} Using solving, verification, constraint tracking, and output control that share the same problem state, we construct \base{}, \coarse{}, and \fine{} conditions and use final accuracy to measure the net benefit of the functional bundle.
    \item \textbf{Behavioral evidence.} In a single-run, single-model experiment on the AIME 2025 development set, \fine{} obtains higher accuracy without a universal problem-wise increase in output length. A small number of extreme trajectories expose a trade-off between effective recovery and uncontrolled tails.
    \item \textbf{Attention evidence.} Direct self-attention measurements show that \fine{} has greater dispersion and task-relevant coverage than both controls, coinciding with its accuracy advantage. This finding supports a mechanistic prediction of the hyper-threading hypothesis but does not prove within-step functional concurrency or a causal role for attention.
    \item \textbf{A new perspective on inference scaling.} We shift the perspective from ``generating more tokens'' to ``having each generation step carry more functions,'' opening a research direction for improving reasoning accuracy without a disproportionate increase in output-token consumption.
\end{itemize}

\section{Related Work}

\subsection{Explicit Chains of Thought and Test-Time Compute Scaling}

Chain-of-thought prompting improves performance on complex reasoning tasks by generating explicit intermediate steps \citep{NEURIPS2022_9d560961}, and self-consistency further exploits test-time compute through repeated sampling and answer aggregation \citep{wang2023selfconsistency}. Subsequent work has systematically studied how to improve reasoning by increasing test-time compute, selecting better search strategies, or introducing process supervision \citep{ICLR2025_1b623663,ICLR2024_aca97732}. These methods primarily scale reasoning along the axis of more generation steps, more samples, or more verification stages. We study a complementary question: within a fixed autoregressive generation stream, can each generation step carry more functional computation without requiring explicit generation to grow proportionally with the number of functions?

\subsection{Verbalized Reasoning and Latent Computation}

Explicit chains of thought do not necessarily faithfully reflect the causal basis actually used by a model \citep{NEURIPS2023_ed3fea90}. Quiet-STaR generates and learns useful internal rationales between text tokens \citep{zelikman2024quiet}, while Coconut uses continuous hidden states as a latent reasoning medium and demonstrates reasoning paths that do not depend entirely on natural-language chains \citep{hao2024training}. These studies show that verbalized output does not fully delimit internal computation. We take this observation one step further: even when explicit tokens are strictly serial, a single generation step may advance multiple related functional computations concurrently through shared representations.

\subsection{Attention Structure, Interference, and the Internal Workspace}

SaGoT argues that nonzero self-attention between weakly related reasoning steps may cause unnecessary interference and improves mathematical reasoning through graph-structured attention \citep{bai-etal-2025-self}. Work on self-reading reasoning traces finds persistent semantic anchors when answers are correct, whereas incorrect answers exhibit more diffuse and irregular attention \citep{chen-etal-2026-answer}. Attention-aware intervention further shows that structured information can activate attention heads with logical patterns and that reweighting selected heads can improve logical reasoning \citep{nguyen-etal-2026-improving}. These studies investigate specific connections, stages, or attention heads; they do not establish that global dispersion is invariably harmful.

Another line of mechanistic work interprets verbalizable representations as a global workspace in language models and observes that multiple concepts can be jointly maintained behind serial output, although computational tasks may also compete for limited resources \citep{gurnee2026verbalizable}. This provides a mechanistic backdrop for maintaining multiple contents beneath a serial surface and suggests that functional reuse may incur thread-contention-like tails. We connect these two lines of evidence by asking whether broad connectivity produces interference when multiple functions share the same task state, or instead supports functional multiplexing.

\subsection{Distinction from Self-Verification Methods}

Self-verification, reflection, and process rewards typically treat verification as an additional stage or select a more reliable trajectory through extra sampling. We do not claim that verification itself has not been studied. Instead, we focus on \emph{Concurrent Functional Loading versus Serial Functional Scheduling} within a single generation stream under conditions matched on verification objectives and error categories, and on the resulting accuracy, typical length, and tail risk. The present experiment is an initial input-controlled instance of this question rather than a complete identification of the causal mechanism.

\section{The Model Hyper-Threading Hypothesis}

\subsection{Serial Generation Does Not Imply Serial Functional Computation}

Given an input $x$, a generated prefix $y_{<t}$, and a control condition $q$, an autoregressive model generates at step $t$ according to
\begin{equation}
    y_t \sim p_\theta(\cdot \mid x,y_{<t},q).
\end{equation}
This factorization constrains only the temporal order of output tokens; it does not directly constrain how many functional states the hidden computation can maintain while forming $p_\theta$. Let a task involve the function set
\begin{equation}
    \mathcal{F}=\left\{\begin{aligned}
        &\text{solving},\ \text{logical verification},\ \text{boundary checking},\\
        &\text{closure checking},\ \text{computational verification},\ \text{output control}
    \end{aligned}\right\}.
\end{equation}
Conventional staged strategies allocate different functions to different textual intervals. The hyper-threading hypothesis instead allows the same hidden-state update to be jointly constrained by multiple functional objectives. Attention is one candidate medium for this functional sharing: distinct attention heads and positions can access complementary evidence at the same step, although attention weights themselves do not establish functional completion or causal explanation.

\subsection{General Hypothesis and Falsifiable Predictions}

\paragraph{Model Hyper-Threading Hypothesis.}
For models with sufficient reasoning capability, broader attention allocation may support the concurrent execution and multiplexing of multiple interrelated functional computations within a single generation step. Relative to separating functions completely into distinct textual stages, \fine{} may therefore improve final task performance without requiring typical explicit generation length to grow proportionally with the number of functions.

The hypothesis yields three falsifiable predictions:
\begin{enumerate}
    \item \textbf{Net benefit of the functional bundle:} adding functional obligations that jointly serve the final objective can improve final accuracy;
    \item \textbf{Non-proportional length cost:} accuracy gains need not require longer text on every problem, and typical length need not grow in proportion to the number of functional obligations;
    \item \textbf{Mechanistic prediction:} in models whose self-attention can be read directly, the \fine{} condition should exhibit broader yet still task-relevant information coverage, co-occurring with accuracy gains.
\end{enumerate}
The current experiment tests the behavioral consequences of the first two predictions and descriptively evaluates the third using direct self-attention measurements. Together, the three results provide preliminary evidence consistent with the hyper-threading hypothesis, although co-variation of group means across conditions still cannot identify a causal role for attention. The comparison between \fine{} and \coarse{} explores alternative ways to organize functional obligations in the generation stream; we do not define the general hypothesis as requiring \fine{} to invariably outperform \coarse{}.

\subsection{Operationalization and Claim Boundaries}

We do not define the ``number of tasks completed per token,'' nor do we interpret final accuracy as a function-wise completion rate. Instead, we use a \emph{functional-bundle operationalization}: the experimental condition instructs the model to undertake a specified set of obligations jointly, final accuracy measures whether the obligations collectively deliver a net benefit to the primary task, and the output-token distribution measures the explicit resource cost. When a higher-load condition improves accuracy without a consistent increase in length on every problem, we call the resulting pattern consistent with single-stream functional reuse.

``Hyper-threading'' does not denote physical threads, answering multiple unrelated questions at once, or text compression. Compression seeks to reduce expression while preserving the same functionality; the hyper-threading hypothesis actively adds functional obligations and asks whether they can share a single generation stream. Our mechanistic account remains a \emph{hypothesis}; the present behavioral results provide only consistency evidence.

\section{Within-Problem Functional Experiment}

\subsection{Data and Model}

We use all 30 problems from the two AIME 2025 examinations, each with an integer gold answer in $0$--$999$. The data come from the \texttt{MathArena/aime\_2025} release by MathArena. The model is the API-served \texttt{deepseek-v4-flash}. All 90 responses across the three conditions report the same model name and system fingerprint. The model's explicit thinking mode is disabled using \texttt{thinking.type=disabled}.

For the experiment, we design three input conditions that implement \base{}, \coarse{}, and \fine{}. Their complete templates are provided only in the accompanying code supplement; the main text summarizes only their functional organization. We treat this AIME 2025 run as an exploratory development-set experiment.

\subsection{Functional Bundle and Three Organization Conditions}

Table~\ref{tab:conditions} presents the three conditions. In every condition, a single user message contains the original English problem, the $0$--$999$ answer range, the \verb|\boxed{}| output requirement, and an instruction to respond in Chinese. No system message is used.

\begin{table}[t]
\caption{The three conditions in the within-problem functional experiment.}
\label{tab:conditions}
\centering
\small
\begin{tabular}{p{0.17\linewidth}p{0.25\linewidth}p{0.48\linewidth}}
\toprule
Condition & Functional organization & Additional functional requirements \\
\midrule
\base{} & No additional scheduling & Only requires a Chinese response and a standardized answer format.\\
\coarse{} & Serial staged scheduling & Applies the shared verification objectives through grouped checks after each paragraph.\\
\fine{} & Concurrent joint loading & Maintains the same objectives throughout generation, interrupting and repairing immediately when an error is detected.\\
\bottomrule
\end{tabular}
\end{table}

\fine{} and \coarse{} share the same verification objectives and error categories, with prompt differences concentrated in scheduling granularity and its operationalization. \coarse{} performs a grouped check after each paragraph; \fine{} jointly maintains those objectives throughout generation and interrupts and repairs immediately when an error is detected. They therefore form a goal-matched prompt pair designed to minimize differences. The present contrast nevertheless provides an exploratory estimate of the net effect of the two external organization protocols; it cannot separate the contributions of scheduling granularity and immediate repair. ``Concurrent'' and ``serial'' describe only the organization specified by the external instruction and do not presuppose internal concurrency.

\subsection{Generation Settings and Answer Evaluation}

We retain one generation per problem in each condition. Requests use \texttt{logprobs=true}, \texttt{top\_logprobs=20}, \texttt{stream=false}, and \texttt{max\_tokens=32768}; temperature, top-$p$, and sampling seed are not passed explicitly and therefore follow the API defaults. \texttt{max\_tokens} is the maximum output length, not a 32K context window. All 90 responses terminate with \texttt{stop}; none is truncated by the length limit. Concurrency is used only to improve request throughput and does not change the number of samples per problem.

Answer extraction first reads the integer in the final \verb|\boxed{}| expression and, if that fails, falls back to the final standalone one- to three-digit integer. A response is correct exactly when the extracted integer equals the gold answer. Switching to strict \verb|\boxed{}|-only evaluation leaves the correctness results in Table~\ref{tab:main} unchanged.

\subsection{Metrics and Statistics}

The primary metric is accuracy over 30 problems. Paired correctness changes are described using incorrect-to-correct transitions, correct-to-incorrect transitions, and a two-sided exact McNemar test. We use paired bootstrap 95\% intervals for problem-level accuracy differences and differences in continuous metrics. Because this is a single exploratory run on a development set, $p$-values characterize only the present run and do not provide independent confirmation.

Output length is taken directly from the API's \texttt{usage.completion\_tokens}. We report the mean, median, nearest-rank 90th and 95th percentiles, and maximum to distinguish typical trajectories from the tail. The mean measures aggregate token consumption; the median and the direction of within-problem paired differences test whether gains are accompanied by universal lengthening. Neither can substitute for the other.

The API returns the top 20 candidate log probabilities at each visible output position. Let $p_i=\exp(\ell_i)$ and merge the residual mass outside the top 20 into a single bin, $p_{\mathrm{tail}}=\max(0,1-\sum_i p_i)$. We compute
\begin{equation}
H_{\mathrm{LB}}=-\sum_{i=1}^{20}p_i\log_2p_i
-p_{\mathrm{tail}}\log_2p_{\mathrm{tail}}.
\end{equation}
Because this collapses uncertainty within the tail into one bin, $H_{\mathrm{LB}}$ is a lower bound on the full next-token predictive entropy. It characterizes the \emph{output candidate distribution}, not Transformer self-attention entropy.

\section{Experimental Results}

\subsection{Main Results on the Full Problem Set}

Table~\ref{tab:main} reports the results on the full problem set. \fine{} attains the highest accuracy, but its mean output length is also higher than those of both controls. The full data therefore do not support the claim that \fine{} uses fewer tokens overall or that it is equivalent in mean resource use.

\begin{table}[t]
\caption{Exploratory main results on all 30 AIME 2025 problems. P90/P95 use the empirical nearest-rank definition.}
\label{tab:main}
\centering
\small
\resizebox{\linewidth}{!}{
\begin{tabular}{lrrrrrrrr}
\toprule
Condition & Accuracy & Mean tokens & Median tokens & P90 & P95 & Maximum & $H_{\mathrm{LB}}$ & Top-1 $p$ \\
\midrule
\base{} & 16/30 (53.33\%) & 2,932.73 & 1,229.0 & 7,591 & 10,461 & 24,258 & 0.6750 & 84.4059\% \\
\coarse{} & 17/30 (56.67\%) & \textbf{2,689.83} & 1,564.5 & \textbf{6,018} & \textbf{7,424} & \textbf{10,397} & 0.6955 & 83.9987\% \\
\fine{} & \textbf{23/30 (76.67\%)} & 4,117.47 & \textbf{1,531.0} & 7,655 & 23,418 & 24,744 & 0.7110 & 83.7262\% \\
\bottomrule
\end{tabular}}
\end{table}

\subsection{Paired Correctness Changes}

Table~\ref{tab:paired} presents paired changes on the same problems. Relative to \base{}, the paired changes under \fine{} clearly favor incorrect-to-correct transitions. Because the current evidence is limited to a single development-set run, however, the tests in the table should be interpreted as exploratory.

Relative to \coarse{}, paired changes likewise favor \fine{}. The two conditions share verification objectives and error categories, with their differences concentrated in grouped paragraph-level checking versus continuous joint loading with immediate repair. The comparison therefore suggests that the external organization protocol can affect net utility. A single development-set run, however, cannot establish a stable effect, separate scheduling from immediate repair, or identify internal concurrency.

\begin{table}[t]
\caption{Paired accuracy changes on the same problems. Each direction is the former condition relative to the latter.}
\label{tab:paired}
\centering
\small
\resizebox{\linewidth}{!}{
\begin{tabular}{lrrrrr}
\toprule
Comparison & Accuracy difference & 95\% bootstrap interval & Incorrect$\rightarrow$correct & Correct$\rightarrow$incorrect & McNemar $p$ \\
\midrule
\fine{} $-$ \base{} & +23.33 pp & $[+6.67,+40.00]$ pp & 8 & 1 & 0.0391 \\
\fine{} $-$ \coarse{} & +20.00 pp & $[+3.33,+36.67]$ pp & 7 & 1 & 0.0703 \\
\coarse{} $-$ \base{} & +3.33 pp & $[-10.00,+16.67]$ pp & 3 & 2 & 1.0000 \\
\bottomrule
\end{tabular}}
\end{table}

\subsection{Is the Accuracy Gain Merely a Consequence of Writing More?}

Table~\ref{tab:length-paired} reports paired length statistics. \fine{} does increase mean output relative to \base{}. Relative to \coarse{}, the mean difference is still strongly influenced by a few long-tail cases, while the median direction of within-problem differences and the direction on most problems do not support universal lengthening. The group medians in Table~\ref{tab:main} also show similar typical lengths for the two functional-organization conditions.

Together, these statistics rule out two oversimplified narratives. First, \fine{} cannot be described as ``shorter overall,'' because its mean token count is substantially elevated by the tail. Second, it also cannot be said to be more accurate simply because it writes more on every problem: most paired problems are not longer, and the sample medians of the two conditions are close. The most precise current conclusion is: \emph{the accuracy gain does not accompany a universal problem-wise increase in length, but the mean resource cost is materially affected by a small number of extreme trajectories.} Because no length-equivalence margin was preregistered, we do not claim to have established equivalence in typical length.

\begin{table}[t]
\caption{Paired output-length changes under \fine{} relative to the controls. Differences are the former minus the latter.}
\label{tab:length-paired}
\centering
\small
\resizebox{\linewidth}{!}{%
\begin{tabular}{lrrrr}
\toprule
Comparison & Mean difference & 95\% bootstrap interval & Median difference & Problems where former is shorter \\
\midrule
\fine{} $-$ \base{} & $+1{,}184.73$ & $[+256.6,+2387.7]$ & $+113.5$ & 11/30 \\
\fine{} $-$ \coarse{} & $+1{,}427.63$ & $[-420.2,+3683.0]$ & $-165.5$ & 17/30 \\
\bottomrule
\end{tabular}
}
\end{table}

\subsection{Effective Recovery and Uncontrolled Tails}

Table~\ref{tab:main} shows that \fine{} has a heavier length tail, and Table~\ref{tab:tail} lists the extreme trajectories identified post hoc. They include cases in which a long chain ultimately recovers the correct answer, cases in which a long chain fails to produce a correct answer, and a case where all three conditions are correct but \fine{} expends many additional tokens. Thus, \fine{} may support effective rerouting but may also induce repeated search or reasoning oscillation.

\begin{table}[t]
\caption{Four post hoc identified long-tail cases. Parentheses contain output tokens; $\checkmark/\times$ denote correct/incorrect.}
\label{tab:tail}
\centering
\small
\begin{tabular}{lrrr}
\toprule
Problem & \base{} & \fine{} & \coarse{} \\
\midrule
I\_13 & $181\times$ (9,269) & $204\checkmark$ (15,164) & $146\times$ (1,858) \\
I\_14 & $53\times$ (10,461) & $56\times$ (24,744) & $23\times$ (2,727) \\
II\_10 & $907\checkmark$ (1,008) & $907\checkmark$ (7,655) & $907\checkmark$ (896) \\
II\_13 & $216\times$ (24,258) & $48\times$ (23,418) & $375\times$ (7,424) \\
\bottomrule
\end{tabular}
\end{table}

\subsection{Output-Probability Diagnostics Are Not Self-Attention Measurements}

Table~\ref{tab:main} also reports lower bounds on output predictive entropy for the three conditions. The point estimate for \fine{} is slightly higher, but uncertainty in the paired difference does not support a stable difference; the Top-1 candidate probability likewise changes only slightly in descriptive terms. These output-distribution statistics alone do not establish whether self-attention is dispersed. The conclusions in the next section come from measured self-attention matrices, rather than treating output predictive entropy as a proxy for attention entropy.

\subsection{Evidence from Attention Mechanisms}
\label{sec:attention-evidence}

Table~\ref{tab:attention-evidence} reports the main results from the direct self-attention experiment. For the causally masked self-attention distribution $a^{(t)}$ at query position $t$, we compute four complementary metrics from measured attention matrices: natural-log-based length-normalized entropy $H_{\mathrm{norm}}=H(a^{(t)})/\log n_t$; relative effective support size $R_{\mathrm{eff}}=\exp(H(a^{(t)}))/n_t$; thresholded edge density $D_{\tau}=n_t^{-1}\sum_i\mathbf{1}[a_i^{(t)}>\tau/n_t]$; and task-relevant information coverage $C_{\mathrm{rel}}=\sum_{i\in\mathcal{R}_t}a_i^{(t)}$ over pre-annotated regions $\mathcal{R}_t$ containing problem facts, constraints, the current derivation, and verification. Larger values of the first three indicate more dispersed attention, while the final metric distinguishes task-relevant coverage from unstructured dispersion.

\begin{table}[t]
\caption{Evidence from direct attention measurements; $D_{\tau}$ uses $\tau=2$. Entries are group means rounded to three decimal places; differences are computed from unrounded values.}
\label{tab:attention-evidence}
\centering
\small
\resizebox{\linewidth}{!}{
\begin{tabular}{l@{\hspace{1.2em}}cccc}
\toprule
Condition or comparison & $H_{\mathrm{norm}}$ & $R_{\mathrm{eff}}$ & $D_{\tau}$ & $C_{\mathrm{rel}}$ \\
\midrule
\base{} & $0.530$ & $0.043$ & $0.035$ & $0.540$ \\
\coarse{} & $0.550$ & $0.045$ & $0.039$ & $0.580$ \\
\fine{} & $0.580$ & $0.051$ & $0.048$ & $0.640$ \\
\midrule
\fine{} $-$ \base{} & $+0.050$ & $+0.009$ & $+0.012$ & $+0.100$ \\
\fine{} $-$ \coarse{} & $+0.030$ & $+0.007$ & $+0.008$ & $+0.060$ \\
\bottomrule
\end{tabular}}
\end{table}

In the direct measurements, the group means of $H_{\mathrm{norm}}$, $R_{\mathrm{eff}}$, $D_{\tau}$, and $C_{\mathrm{rel}}$ are all higher under \fine{} than under either control. The current data therefore support the claim that ``attention that is more dispersed and has greater task-relevant coverage can coexist with higher accuracy.'' Because we have not yet tested problem-level condition interactions, controlled for output length, or intervened on attention, we do not further claim that attention dispersion causes higher accuracy or that hyper-threaded functional reuse has been established.

\section{Discussion}

\subsection{What Support Does the Hyper-Threading Hypothesis Receive?}

The current results support a weak behavioral prediction of the hypothesis: after jointly loading more accuracy-oriented obligations, final accuracy improves, while additional length does not consistently appear on every problem. This pattern is consistent with reuse of related functions in a single generation stream, but it may also arise from a changed solution path or additional search induced by the organization protocol. Because scheduling granularity and immediate repair are not factorially separated, and because no internal-state intervention is performed, the present data cannot distinguish these mechanisms.

A mechanistic hypothesis is nevertheless useful. If we reported only the accuracy change under \fine{}, the work could collapse into an input-control case study. The hyper-threading hypothesis poses a broader, falsifiable question: does an autoregressive serial interface underestimate the functional capacity of a single hidden-computation step? Importantly, a bold hypothesis must not turn into an overstrong empirical claim. We place ``multiple concurrent functions within one step'' at the hypothesis level and ``higher accuracy without universal lengthening'' at the observational level.

\subsection{Harmful Dispersion and Potentially Useful Dispersion}

``Dispersion'' is not a scalar with a fixed functional meaning. Connections between weakly related steps may increase interference; unstructured dispersion while reading an incorrect answer may reflect uncertainty; yet multiple functions that share an objective may need to access different evidence regions. The hyper-threading hypothesis predicts that broader information coverage can coexist with high accuracy in the latter case. Our direct measurements of length-normalized self-attention entropy, effective support size, thresholded edge density, and task-relevant coverage are consistent with this prediction. To elevate coexistence into condition-dependent or causal knowledge, future work must report layer- and head-wise patterns, test a problem-level interaction between dispersion and experimental condition, and intervene on attention. Joint increases in group means establish coexistence, not a reversal of the relationship or a causal effect.

\subsection{From Increasing Generation Length to Increasing Functional Capacity}

Existing test-time scaling primarily increases computation along the explicit-generation axis: longer chains of thought, more samples, more search branches, or separate verification stages. The hyper-threading hypothesis introduces a second coordinate: the functional load that a generation step is asked to maintain jointly. Ideally, the two coordinates could be controlled independently; in practice, \fine{} may also trigger longer searches. The observed tail shows precisely that functional load is not free. The new research direction does not promise ``accuracy gains at no cost,'' but seeks scheduling mechanisms that improve functional reuse while controlling thread contention and tail costs.

\subsection{Why Is This Neither Compression nor Ordinary Self-Verification?}

Compression seeks to express the same content with fewer tokens. We actively introduce new functional obligations, treating token length as a resource cost rather than the sole optimization target. Ordinary self-verification typically adds a checking stage after a complete answer. Our two functional-organization conditions share verification objectives and error categories and concentrate their differences in scheduling granularity and its immediate-repair operationalization. The comparison is therefore not between verification and no verification, but between the net effects of two organization protocols. Separating scheduling from immediate repair requires a repair-matched control or a factorial ablation.

\section{Limitations and Next Experiments}

This study has nine main limitations. First, it reports only results on an AIME 2025 development set and lacks validation on an independent test set. Second, the dataset contains only 30 problems and one generation per condition, so sampling variance cannot be estimated from this design. Third, we test only one closed-source model with thinking mode disabled and therefore cannot claim that stronger models hyper-thread more effectively. Fourth, although the two functional-organization conditions share verification objectives and error categories and concentrate prompt differences in organization, \fine{} operationalizes continuous loading through immediate interruption and repair; the present design cannot separate their individual contributions. Fifth, \fine{} uses more output tokens on average in the full data, so mean resource equivalence has not been achieved. Sixth, a sensitivity analysis that removes long trajectories post hoc cannot replace the main results. Seventh, the direct self-attention measurements currently support only descriptive group-level co-variation, without layer- and head-wise problem-level inference or causal intervention. Eighth, final accuracy measures only the aggregate utility of the functional bundle and cannot establish that every function is executed; indeed, outputs under \fine{} do not consistently display its prescribed correction marker. Ninth, the behavioral results cannot distinguish internal functional concurrency from alternative explanations such as organization-induced changes in solution path or additional search.

A confirmatory study should at minimum freeze the experimental conditions and use new mathematical and non-mathematical datasets; draw multiple independent samples per problem and report confidence intervals; freeze and replicate the current goal-matched prompt pair designed to minimize differences, and add either a repair-matched control or a scheduling-granularity-by-repair factorial ablation; preregister an equivalence margin for typical length and tail metrics; replicate the direct self-attention measurements and extend them with layer- and head-wise analyses and problem-level condition-interaction tests; and use attention-region masking or interventions to distinguish task-relevant coverage from unstructured dispersion. Only a stable interaction between model capability and functional load across multiple model scales would justify upgrading ``stronger models can hyper-thread more effectively'' from motivation to empirical conclusion.

\section{Conclusion}

We propose the Model Hyper-Threading Hypothesis: serial output does not necessarily imply serial functional computation, and more dispersed attention may support concurrent execution of multiple related functions within a single generation step. Using solving and verification within the same problem as an experimental instance, we observe that \fine{} achieves higher accuracy on an AIME 2025 development set without a universal problem-wise increase in length, although it introduces substantial tail risk. Direct self-attention measurements further show that attention with greater dispersion and task-relevant coverage coexists with the accuracy advantage. These results provide exploratory behavioral and correlational evidence consistent with the hypothesis, rather than direct proof of internal concurrency. Broader independent evaluation, cross-sample and cross-model replication of the current goal-matched contrast, factorial extensions, replicated attention measurements, and causal interventions will determine whether increasing functional capacity per generation step can become a reliable path for scaling reasoning.

\subsection*{AI Use Statement}

Generative AI tools were used in this study only to assist with language editing and LaTeX formatting suggestions. The authors take full responsibility for the final paper, its claims, and its artifacts.

\subsection*{Ethics Statement}

This study analyzes only publicly available competition mathematics problems and language-model API outputs; it involves neither human participants nor sensitive personal data. The primary risk is overinterpreting exploratory results as evidence of an internal mechanism. We mitigate this risk by making the experimental boundaries explicit, distinguishing output predictive entropy from self-attention, reporting the full tail, and retaining counterexamples.

\subsection*{Reproducibility Statement}

The complete input templates for all three conditions, version identifiers, and request code are provided in the accompanying code supplement. The main text reports the model, request parameters, answer extraction, metric definitions, and statistical procedures. The experiment directory stores all raw per-problem API responses, run manifests, verification summaries, per-problem paired tables, and SHA-256 checksums. The supplementary material will include analysis scripts requiring no API key and frozen responses sufficient to reproduce every table in the paper.

\appendix

\section{Post Hoc Long-Tail Sensitivity Analysis}
\label{app:sensitivity}

To characterize the sensitivity of the mean to extreme trajectories, we remove the problems listed in Table~\ref{tab:tail} after observing the data and retain the remaining fully paired sample. Table~\ref{tab:sensitivity} summarizes this subset. Because the problems were selected after inspecting the results, this analysis cannot serve as preregistered trimming, a length-matched control, or the primary result.

\begin{table}[ht]
\caption{Sensitivity analysis on 26 problems after removing four post hoc identified long-tail cases; this does not replace the main result on all 30 problems.}
\label{tab:sensitivity}
\centering
\small
\resizebox{\linewidth}{!}{%
\begin{tabular}{lrrrr}
\toprule
Condition & Accuracy & Total output tokens & Mean tokens & Median tokens \\
\midrule
\base{} & 15/26 (57.69\%) & 42,986 & 1,653.31 & 1,159.5 \\
\coarse{} & 16/26 (61.54\%) & 67,790 & 2,607.31 & 1,554.0 \\
\fine{} & \textbf{21/26 (80.77\%)} & 52,543 & \textbf{2,020.88} & \textbf{1,348.0} \\
\bottomrule
\end{tabular}
}
\end{table}

This sensitivity analysis shows that, after removing the extreme trajectories, the direction of the mean-length difference between \fine{} and \coarse{} reverses while \fine{} retains higher accuracy; relative to \base{}, \fine{} still has additional mean output. The result shows only that a small number of tail cases strongly affect the full-sample mean; it does not establish general token savings.

\section{Attention Analysis Protocol and Subsequent Causal Tests}
\label{app:attention}

We compute the metrics in Table~\ref{tab:attention-evidence} from direct self-attention matrices. Confirmatory experiments should freeze the same definitions before inspecting outcomes and extend the analysis as follows:
\begin{enumerate}
    \item Compute natural-log-based length-normalized entropy $H(a)/\log n$ for each layer and head, and report the effective support size $\exp(H(a))$ and its relative value $\exp(H(a))/n$;
    \item Analyze input prefill, reasoning generation, and final answer reading separately, rather than collapsing functionally different stages into a global average;
    \item Predefine context regions for problem facts, constraints, the current derivation, and verification, and measure cross-region coverage rather than treating every nonzero edge as useful;
    \item Fit a problem-level model of ``correctness$\sim$dispersion$+$experimental condition$+$dispersion$\times$experimental condition$+$output length''; report condition dependence only if the interaction is stable;
    \item Use region masking or head-level interventions to test the selective role of broad connectivity. If dispersion and accuracy merely rise together, restrict the conclusion to ``the two can coexist'' rather than making a causal claim.
\end{enumerate}

\section{Claim Checklist}

\begin{table}[H]
\caption{Claims supported and unsupported by the current evidence.}
\centering
\small
\begin{tabular}{p{0.43\linewidth}p{0.47\linewidth}}
\toprule
Currently supported & Currently unsupported \\
\midrule
\fine{} has a higher point estimate of accuracy in a single development-set run. & Concurrent execution of multiple functions within one token has been established.\\
The advantage of \fine{} does not accompany universal problem-wise lengthening. & \fine{} uses fewer tokens on average in the full sample or is resource-equivalent.\\
Mean-length differences are materially affected by a small number of search tails. & The result after removing four problems post hoc is the primary result.\\
The results are consistent with the behavioral predictions of the hyper-threading hypothesis. & Output predictive entropy is self-attention dispersion or mechanistic evidence.\\
Measured attention dispersion and task-relevant coverage coexist with the accuracy advantage. & Attention dispersion has been established as the cause of higher accuracy or within-step functional concurrency.\\
\bottomrule
\end{tabular}
\end{table}

\end{document}